\documentclass[10pt,twocolumn,table,letterpaper]{article}
\usepackage[svgnames,table]{xcolor} 
\usepackage{cvpr}
\usepackage{times}
\usepackage{epsfig}
\usepackage{graphicx}
\usepackage{amsmath}
\usepackage{amssymb}
\usepackage{marvosym}
\usepackage{subfigure}
\usepackage{multirow,bigstrut}
\usepackage{algorithm}
\usepackage{algorithmic}
\usepackage{dsfont}
\usepackage{flushend}
\usepackage{booktabs}

\usepackage[pagebackref=true,breaklinks=true,letterpaper=true,colorlinks,bookmarks=false]{hyperref}

\cvprfinalcopy 

\def\cvprPaperID{237} 

\newcommand*{\titleAT}{\begingroup 
\newlength{\drop} 
\drop=0.05\textheight 

\includegraphics[scale=1.5]{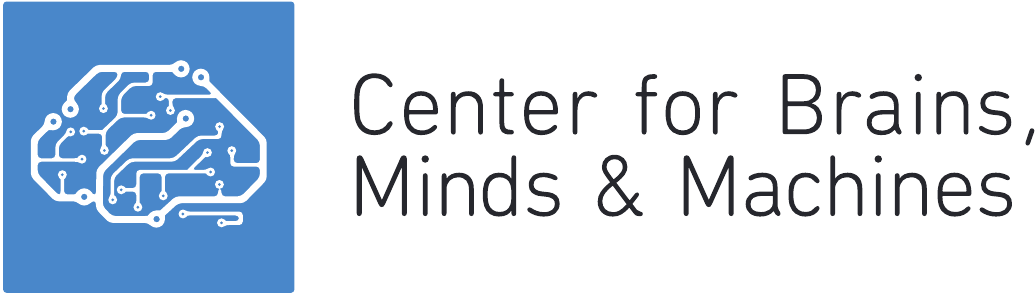}

\textcolor{CornflowerBlue}{\rule{\textwidth}{3 pt}}\par 
\vspace{2pt}\vspace{-\baselineskip} 
\rule{\textwidth}{0.4pt}\par 

\vspace{\drop} 
\textbf{\textsf{\large{CBMM Memo No. \memonumber}}}\quad \quad \quad\quad \quad \quad \quad\quad\quad \quad\quad\quad      \textbf{\large{\memodate}}

\vspace{\drop}
\begin{center}
\textbf{\textsf{\huge{\memotitle}}}\\
\vspace{0.4\drop}
\textbf{\Large{\textsf{by}}}\\
\vspace{0.4\drop}
\textbf{\textsf{\large{\memoauthors}}}
\end{center}
\vspace{\drop}
\textbf{\textsf{\large{\noindent Abstract}:}} {\memoabstract}

\textcolor{CornflowerBlue}{\rule{\textwidth}{3 pt}}\par 
\vspace{2pt}\vspace{-\baselineskip} 
\rule{\textwidth}{0.4pt}\par

\begin{minipage}{.15\linewidth}
\includegraphics[scale=0.1]{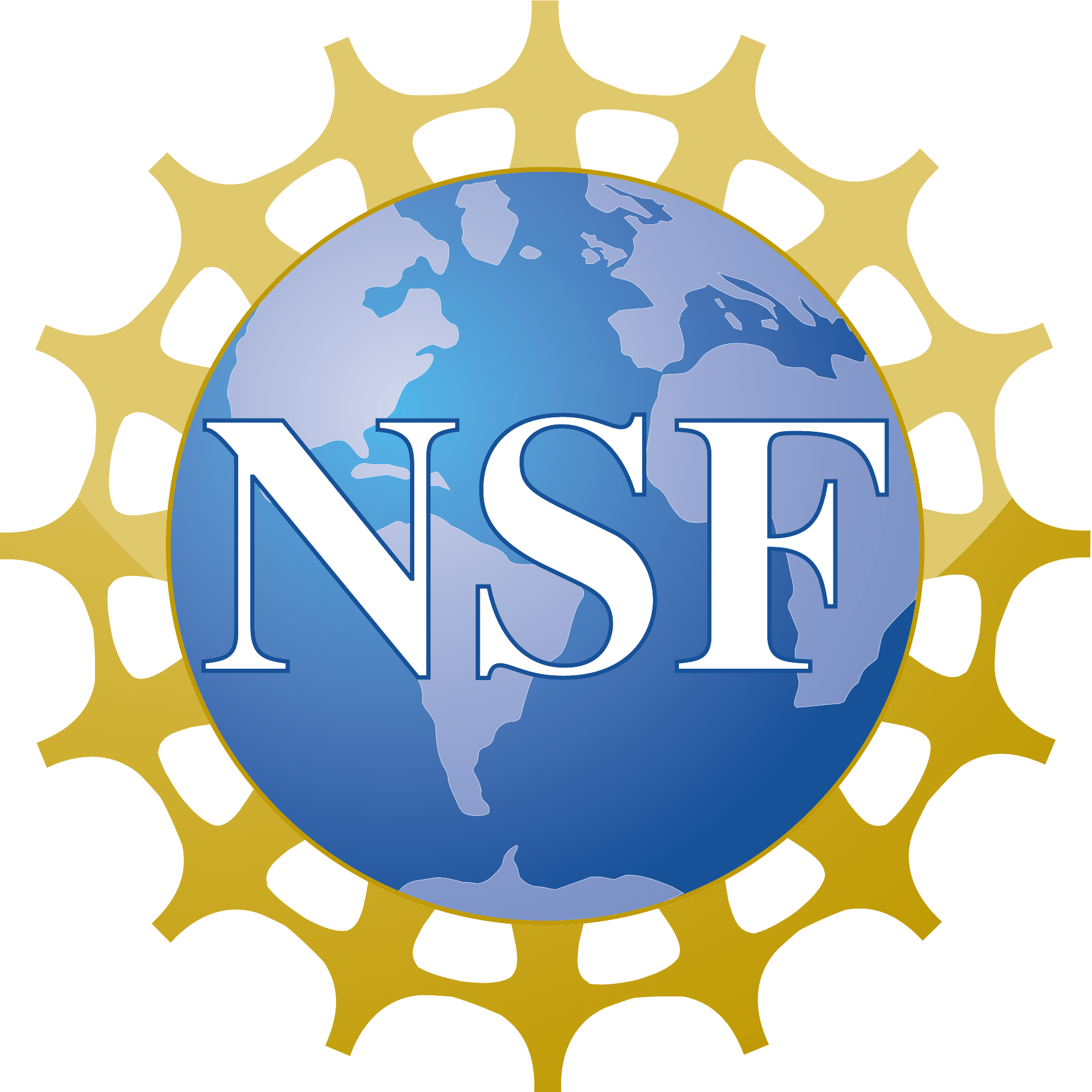}
\end{minipage}
\begin{minipage}{.84\linewidth}
\textbf{\textsf{\large{This material is based upon work supported by the Center for Minds, Brains and Machines (CBMM), funded by NSF STC award CCF-1231216.}}}
\end{minipage}
\endgroup}

\author{
Xianjie Chen$^{1}$,
Roozbeh Mottaghi$^{2}$,
Xiaobai Liu$^{1}$,
Sanja Fidler$^{3}$,
Raquel Urtasun$^{3}$,
Alan Yuille$^{1}$\\
$^{1}$University of California, Los Angeles~~
$^{2}$Stanford University~~
$^{3}$University of Toronto \\
{\tt\small \{cxj,lxb\}@ucla.edu~roozbeh@cs.stanford.edu~\{fidler,urtasun\}@cs.toronto.edu~yuille@stat.ucla.edu}
}

\begin{document}

\thispagestyle{empty}

\onecolumn

\def\memonumber{ \textsf{015}} 
\def\memodate{\textsf{\today}} 
\def\memotitle{\textsf{Detect What You Can: Detecting and Representing Objects using Holistic Models and Body Parts}} 
\def\memoauthors{\textsf{\mbox{Xianjie Chen$^{1}$,Roozbeh Mottaghi$^{2}$,Xiaobai Liu$^{1}$,Sanja Fidler$^{3}$,Raquel Urtasun$^{3}$,Alan Yuille$^{1}$}}\\
$^{1}$University of California, Los Angeles~~
$^{2}$Stanford University~~
$^{3}$University of Toronto \\
{\tt\small \{cxj,lxb\}@ucla.edu~roozbeh@cs.stanford.edu~\{fidler,urtasun\}@cs.toronto.edu~yuille@stat.ucla.edu}}

\def\memoabstract{\textsf{Detecting objects becomes difficult when we need to deal with large shape deformation, occlusion and low resolution. We propose a novel approach to i) handle large deformations and partial occlusions in animals (as examples of highly deformable objects), ii) describe them in terms of body parts, and iii) detect them when their body parts are hard to detect (\eg, animals depicted at low resolution). We represent the holistic object and body parts separately and use a fully connected model to arrange templates for the holistic object and body parts. Our model automatically decouples the holistic object or body parts from the model when they are hard to detect. This enables us to represent a large number of holistic object and body part combinations to better deal with different ``detectability'' patterns caused by deformations, occlusion and/or low resolution. We apply our method to the six animal categories in the PASCAL VOC dataset and show that our method significantly improves state-of-the-art (by 4.1\% AP) and provides a richer representation for objects. During training we use annotations for body parts (\eg, head, torso, etc), making use of a new dataset of fully annotated object parts for PASCAL VOC 2010, which provides a mask for each part. }}

\titleAT
\newpage
\twocolumn
\title{Detect What You Can: Detecting and Representing Objects using Holistic Models and Body Parts}
\maketitle

\begin{abstract}
Detecting objects becomes difficult when we need to deal with large shape deformation, occlusion and low resolution. We propose a novel approach to i) handle large deformations and partial occlusions in animals (as examples of highly deformable objects), ii) describe them in terms of body parts, and iii) detect them when their body parts are hard to detect (\eg, animals depicted at low resolution). We represent the holistic object and body parts separately and use a fully connected model to arrange templates for the holistic object and body parts. Our model automatically decouples the holistic object or body parts from the model when they are hard to detect. This enables us to represent a large number of holistic object and body part combinations to better deal with different ``detectability'' patterns caused by deformations, occlusion and/or low resolution. 

We apply our method to the six animal categories in the PASCAL VOC dataset and show that our method significantly improves state-of-the-art (by 4.1\% AP) and provides a richer representation for objects. During training we use annotations for body parts (\eg, head, torso, etc), making use of a new dataset of fully annotated object parts for PASCAL VOC 2010, which provides a mask for each part. 
\end{abstract}

\section{Introduction}
\begin{figure}
\centering
  \includegraphics[width=20pc]{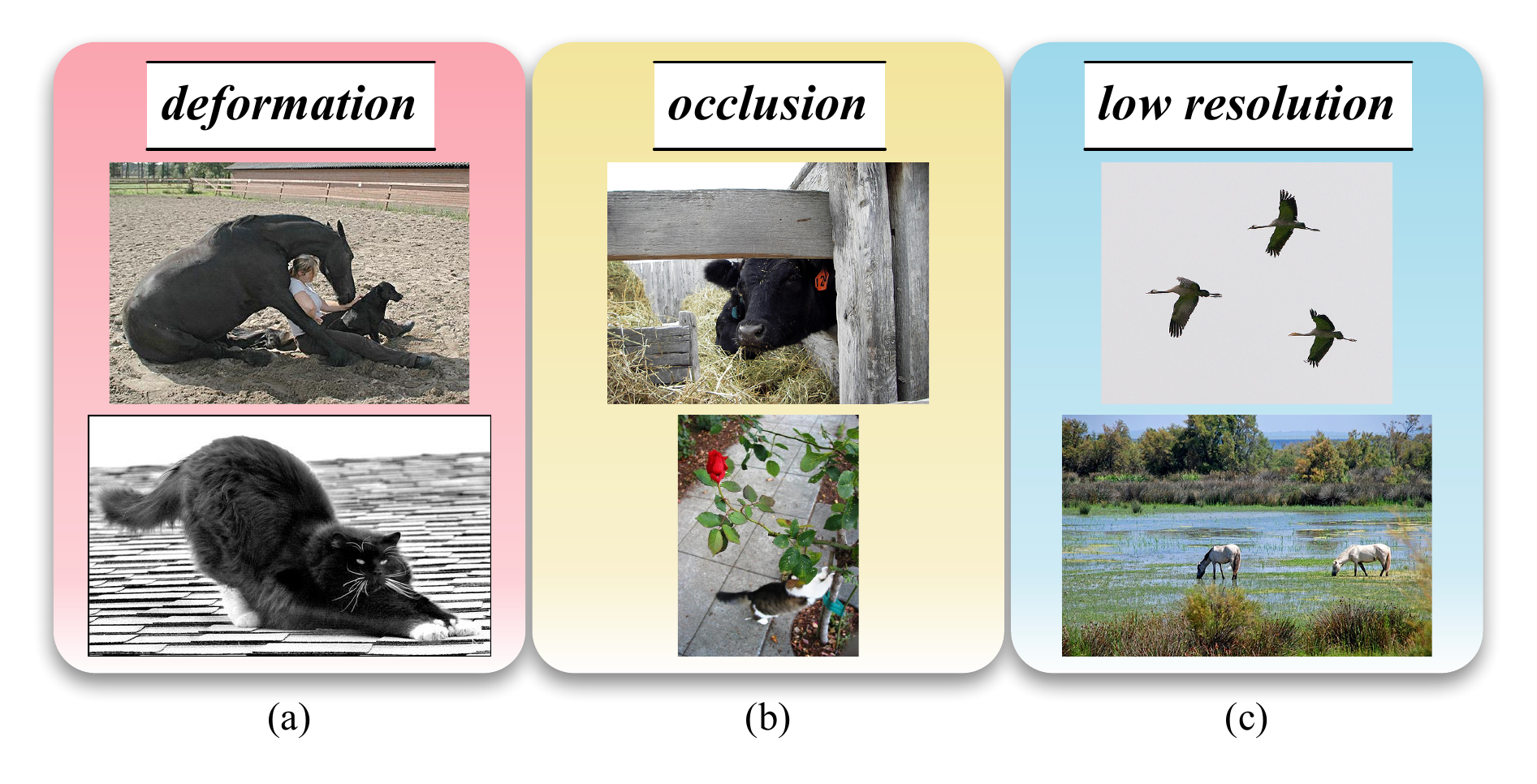}
  \caption{Handling deformation, occlusion, and low resolution objects are challenging tasks for object detectors. {\bf (a)} It is hard to detect the objects due to high deformation, but some of the body parts (\eg, head) can be detected reliably. {\bf (b)} Detecting the occluded body parts is difficult. {\bf (c)} It is possible to detect the objects but not their parts (\eg, the low resolution head). Our strategy is to detect what we can -- body parts or the holistic object -- by using a model that adaptively switches off some of the parts (or the holistic object) in the model. }
  \label{fig:motiv}
\end{figure}

Despite much recent progress in detecting objects, dealing with large shape deformations, 
occlusions and low resolution remain fundamental challenges. This is evident from the low performance of state-of-the-art object detectors~\cite{felzenswalb10, zhu10,vedaldi09} for animals, which are highly deformable objects.
Figure~\ref{fig:motiv} shows some typical examples of difficult images for detecting animals: (a) Animals can be highly deformed, in which case, finding a template for the holistic object (`root') becomes hard due to deformation. However, some of the constituent body parts (\eg, head or torso) may still be reliably detected. Hence, the detector should rely on those parts to detect the holistic object. (b) Some body parts of the animals might not be detectable due to occlusion, truncation or local ambiguity. Therefore, the object detector should automatically select a subset of reliable body parts for detection of the holistic object. (c) The visual cues for body parts are typically quite weak when an object is small so the body parts should be ignored in the model.

In this paper, we propose a novel method to better detect objects in the scenarios mentioned above. We introduce a flexible model that decouples body parts or the holistic object from the model when they cannot be detected hence we ``detect what we can". There are two main advantages to our model: 1) The model provides flexibility by allowing the holistic object (as well as the body parts) to be ignored. The current detectors typically model objects by templates for the holistic object and a set of parts (\eg, \cite{felzenswalb10} or~\cite{azizpour12}, which allows missing body parts) or just a set of templates for parts (\eg, \cite{yang11}). 2) We use `detectability' to model occlusion and deformation and not `visibility'. For example, the holistic body of a highly deformed cat is visible, but we would rather not model it since it is not easily `detectable'.

We use a fully connected graphical model where the nodes represent the holistic object and body parts. The edges encode spatial and scale relationships among the parts and between the parts and the holistic object.  Our
model contains switch variables for each node, which enable us to decouple the nodes from the model if the corresponding body parts, or the holistic object cannot be detected. These switch variables give our model the ability to represent a large number of combinations of holistic object and body parts (Fig.~\ref{fig:model}). We perform inference efficiently on our fully connected model by taking advantage of the fact that the nodes are shared between different detectability patterns. 

We perform an extensive set of experiments to compare our method with DPM~\cite{felzenswalb10} and segDPM~\cite{fidler13}. We show that our model provides significant improvement over DPM~\cite{felzenswalb10} and segDPM~\cite{fidler13} (7.3\% and 4.1\% AP, respectively), while it provides a richer description for objects in terms of body parts (compared to a simple bounding box). We also quantify our method's ability to localize body parts. We provide a new dataset of annotated body parts for all of the animal categories of PASCAL VOC 2010, which is used for training and evaluating our model. Animals in PASCAL are highly deformable and appear at different scales with different degree of occlusion, hence provide a suitable testbed for our method.



\section{Related Work}


There is a considerable body of work on part-based object detectors. Some of these methods represent an object with a holistic object (``root") template attached to a fixed number of parts~\cite{felzenswalb10,zhu10,chen2010active}. The main disadvantage of these methods is that they are not robust against occlusion. Azizpour et al.~\cite{azizpour12} propose a DPM that allows missing parts. Our method is different from theirs as we consider \emph{detectability} of parts instead of \emph{visibility}, and our model is more flexible as we can switch off the ``root" as well. We also show significantly better results. Girshick et al.~\cite{girshick11} propose a grammar model to handle a variable number of parts. The difference between our method and theirs is that we consider occlusion of body parts, while they model occlusion for latent parts of the model. In~\cite{park10}, the authors propose a multi-resolution model to better detect small objects, where the parts are switched off at small scales. In contrast, 
we do not explicitly incorporate size into the model and let the model choose whether the parts are useful to describe an object or not. 

The recent works on human pose estimation~\cite{yang11,johnson2010clustered,Singh10,Sapp10} typically do not consider the holistic object (root) template in their model as they are faced with highly variable poses. Since detecting the individual parts can be hard when objects are small, having a root model can be helpful so our model adaptively switches on or off the root.



In contrast to object detectors encoding parts using latent variables~\cite{Crandall06, zhu08, fidler09,dollar08, felzenswalb10, Schnitzspan10, Karlinsky10}, we use semantic body parts. Most of these methods formulate object detection within a discriminative framework and so learn parts that are most effective for discriminating the object from the background in images. Hence many of these methods are not suitable for recovering the body parts of the object. 

Strong supervision has been used in various ways to detect objects~\cite{Bourdev09, sun11, Branson11, parkhi11, azizpour12,gu12}. 
Similar to our method, \cite{azizpour12, sun11, Branson11} use additional annotations to provide supervision for a part-based model, while Poselets~\cite{Bourdev09} and \cite{gu12} use keypoints and masks annotations to cluster objects or their parts. \cite{parkhi11} use head as a distinctive part to detect cats and dogs with the assumption that the appearance of animals is homogeneous.


\section{Model}
\label{sec:model}
\begin{figure*}
\centering
  \includegraphics[width=0.995\linewidth]{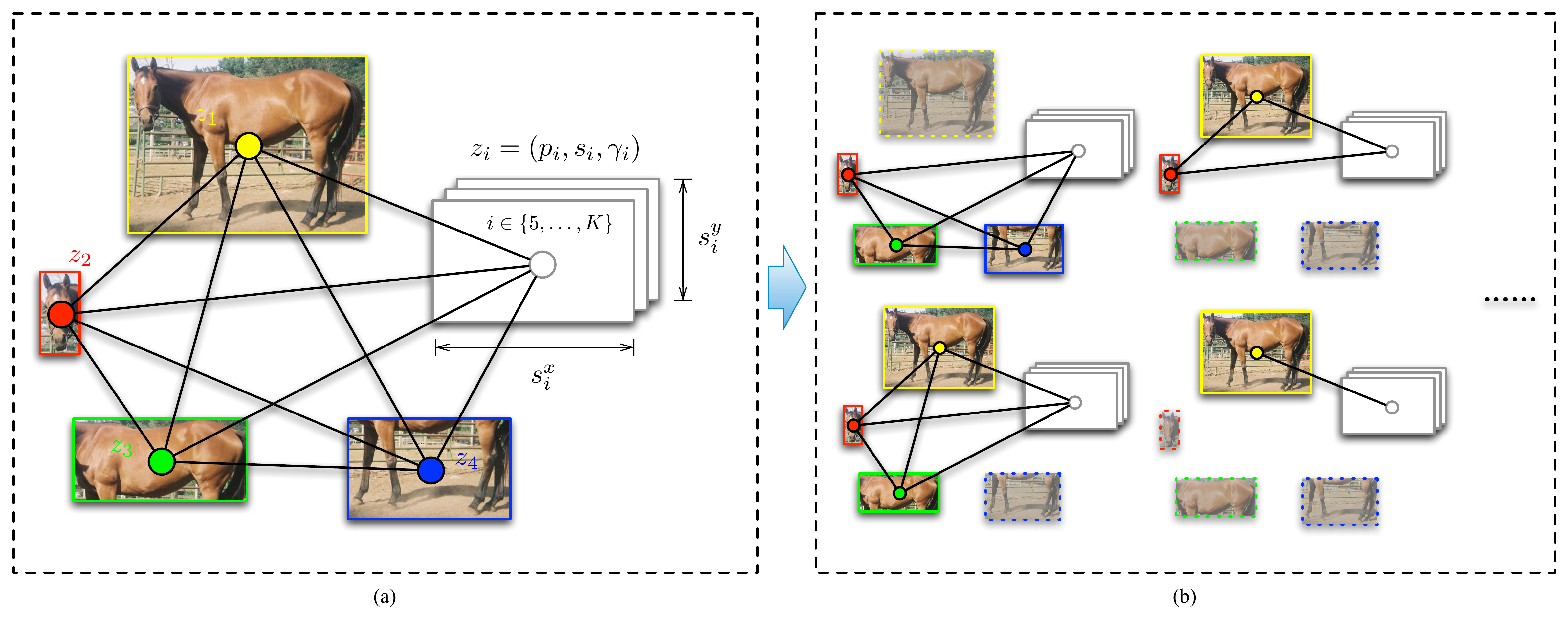}
  \vspace{-2.5mm}
  \caption{{\bf (a)} Our graphical model where the nodes represent the holistic object and its body parts. Their state variables are position, scale, and switch variable ($\gamma$). The holistic object is shown in yellow and some example body parts are shown in red, green, and blue. The rest of the body parts are shown with white rectangles. {\bf (b)} The switch variables decouple nodes from the graph, depending on which parts are detected, and enable the model to deal with different detectability patterns. Boxes with dashed border are those decoupled from the graph.}  \label{fig:model}
  \vspace{-2.5mm}
\end{figure*}

Our overall goal is to better handle large shape deformation, occlusion and low resolution by introducing a flexible structure that allows part and/or holistic object (root) templates to be switched on or off. 

We represent objects by their holistic model and a collection of body parts. We arrange them in a fully connected graph as shown in Figure~\ref{fig:model}. We consider connections between all pairs of nodes because if we use a structure like a tree,  switching off non-leaf nodes will break the structure. The connection between pairs of nodes characterizes their spatial and scale relationships.

Formally, a random variable $z_i$ is associated to node $i$ and represents its location $p_i$, size $s_i$, and switch variable $\gamma_i$. Here $i \in \{1, \dots, K \}$, $p_i \in \{1, \dots, L_i\}$, $s_i \in \{1, \dots,S_i\}$ and $\gamma_i \in \{0, 1\}$, where $K$ is the number of all nodes in the graph, and $L_i$ and $S_i$ are the possible number of positions and scales of node $i$, respectively. Here $\gamma_i=0$ indicates node $i$ should be switched off, which means it is hard to detect part $i$. We denote this $K$-node relational graph as $G=(V,E)$, where the edges specify the spatial and scale consistency relations. For notational convenience, we use the lack of subscript to indicate a set spanned by the subscript (\eg, $\gamma=\{\gamma_1, \dots, \gamma_K \}$).

The score associated with a configuration of node locations $p$, sizes $s$ and switch variables $\gamma$ in image $\mathbf{I}$ can be written as follows:
\begin{equation}
\label{eq:score}
F({\bf z}) = \sum_{i \in V} w_i \phi(\mathbf{I},z_i) + \sum_{ij \in E} \boldsymbol{w_{ij}} \boldsymbol\psi(z_i, z_j) + b(\gamma),
\end{equation}
where $\phi(.)$ is the appearance term and $\boldsymbol\psi(.,.)$ denotes the pairwise relationships. The pairwise relationship between nodes consists of spatial $\boldsymbol\psi_{sp}$ and scale $\boldsymbol\psi_{sc}$ terms. The last term, $b(\gamma)$, is a scalar bias term that models the prior of detectability pattern $\gamma = \{\gamma_1, \dots, \gamma_K\}$. When a node is switched off, we do not model its appearance or relationship with others. In particular, we have:
\begin{eqnarray*}
\phi(\mathbf{I},z_i) &=& \begin{cases} \phi(\mathbf{I},p_i, s_i) & \mbox{if } \gamma_i \equiv 1 \\
0 & \mbox{if } \gamma_i \equiv 0. \end{cases} \\
\boldsymbol\psi(z_i, z_j) &=& \begin{cases} \Big[\boldsymbol\psi_{sp}(p_i,p_j)\,\,\,\boldsymbol\psi_{sc}(s_i,s_j)\Big]^T & \mbox{if } \gamma_i  \gamma_j \equiv 1\\
\boldsymbol{0} & \mbox{otherwise}. \end{cases}
\end{eqnarray*}
Here $\phi(\mathbf{I},p_i, s_i)$ models the appearance of node $i$ inside a window defined by $p_i$ and $s_i$. There are various choices to build the appearance model. In our experiments, we train the appearance model of each node separately using \cite{felzenswalb10} and \cite{fidler13}. See more details in Section~\ref{sec:exp}. 



\begin{figure}[t!]
\centering
  \includegraphics[width=0.88\linewidth,trim=10 12 10 10,clip=true]{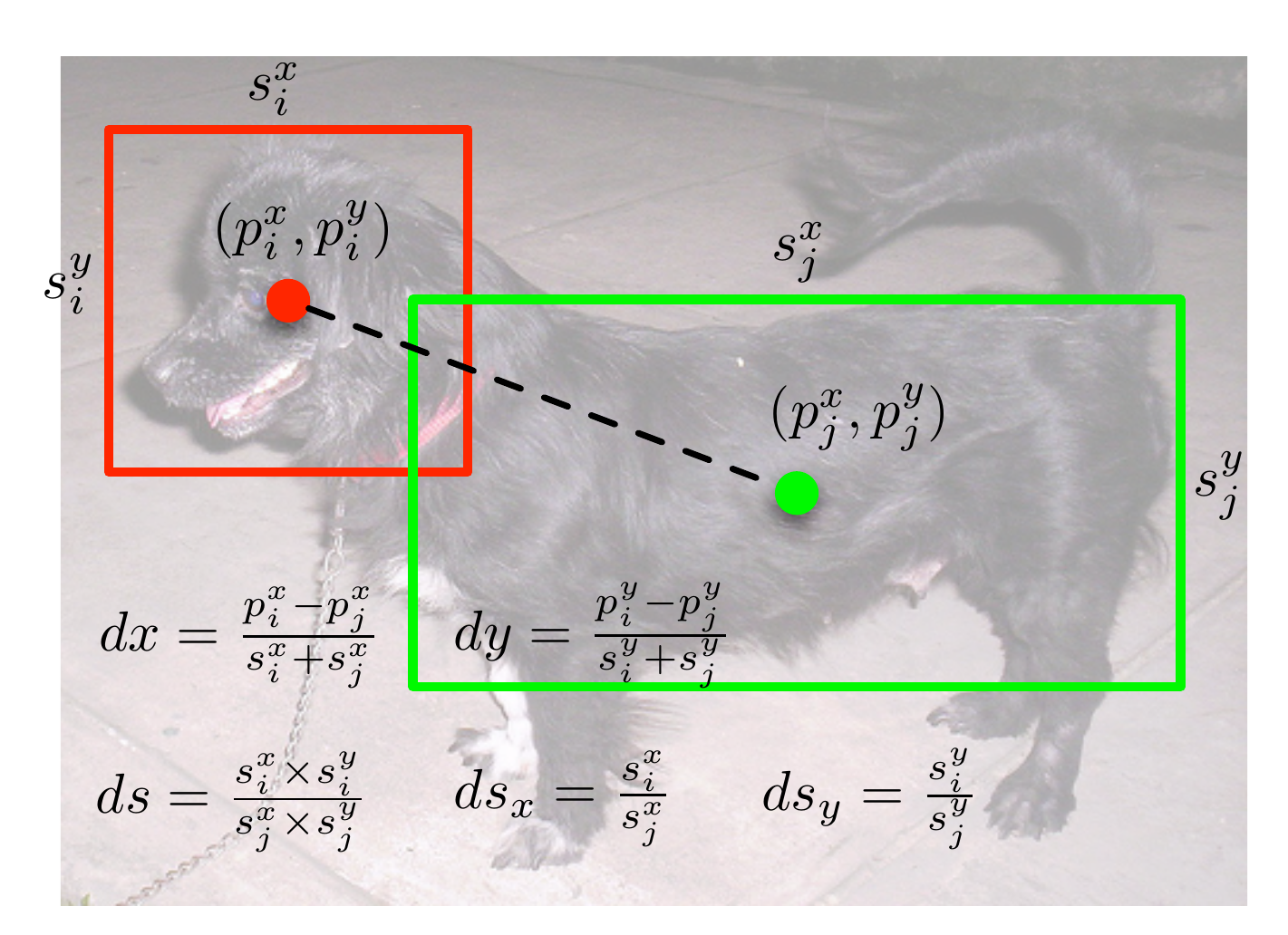}
  \vspace{-0mm}
  \caption{The deformation and scale features used in our model.} 
  \vspace{-2mm} 
  \label{fig:spatial}
\end{figure}

We use similar spatial deformations as used by~\cite{felzenswalb10}: $\boldsymbol\psi_{sp}(p_i, p_j) =  \left[dx, dy, dx^2, dy^2\right]$. The difference is that we normalize the distances by the sum of the size of the involved nodes. The scale term for two nodes is defined by $\boldsymbol\psi_{sc}(s_i, s_j) =  \left[ds, ds_x, ds_y, ds ^ 2, ds_x^ 2, ds_y ^ 2\right]$, where:
\begin{equation}
ds = \frac{s_i^x \times s_i^y}{s_j^x \times s_j^y}, \,\,ds_x = \frac{s_i^x}{s_j^x}, \,\,ds_y = \frac{s_i^y}{s_j^y}
\end{equation}
These terms are visualized in Fig.~\ref{fig:spatial} for clarity.


By introducing the switch variable for each node, our model represents a large number of holistic object and body parts combinations that are suitable for different detectability patterns. The intuition for having the switch variables is as follows. For example, in some cases, it is hard to find the body parts of an object, and a model for the holistic object works better than a model with body parts. This case typically happens for objects with low resolution, where it is hard to detect the body parts due to the lack of appearance information. Or, often it is hard to model the appearance of a highly deformed object. However, the body parts might be detected reliably. Thus, models that only capture the relationship between the body parts and do not rely on the holistic object are more suitable. Additionally, models that switch off a subset of body parts are suitable to detect partially occluded objects.

\subsection{Inference}
To detect the optimal configuration for each object, we search for the configurations of the random variables $z$ that maximize our score function ${\bf z}^* = \arg \max_{{\bf z}}F({\bf z})$. The output configuration ${ {\bf z}^* = ({\bf p}^*, {\bf s}^*, \mathbf\gamma^*)}$ indicates whether the body parts or the holistic object are detected and specifies their corresponding positions and scales. Score $F({\bf z}^*)$ is used as a measure of confidence for the detected object. 

Note that our model is loopy and the space of possible positions and scales for each node is large. Therefore, we adopt a procedure that results in an approximate solution and performs reasonably well for this problem. 

We use the appearance term $\phi$ to reduce the search space of positions and scales. In particular, for each node, we set a threshold on its appearance term $\phi$ and perform spatial non-maximum suppression to select a limited number of hypotheses for that node. Similarly, \cite{johnson2010clustered} used a cascaded method to generate part hypotheses for human pose estimation. The thresholding might hurt their approach because if a part is missed as the result of thresholding, it will not be possible to recover it in later stages. In contrast, our model allows parts to be switched off and thus is robust to missing parts, and a limited number of strong hypotheses is sufficient for our model.

Although the number of possible state configurations grows exponentially with the number of nodes in the graph, the number of body parts is small in our case (\eg, head, torso, and legs). In our experiments, the number of hypotheses for each node is typically around 15 per image. Therefore, we need to evaluate at most a few thousands of configurations for each image, which is certainly feasible by exhaustive search. Note that the nodes are \emph{shared} between different detectability patterns so we do not need to perform inference separately for each pattern. 

\subsection{Learning}
\label{se:learn}

To learn the parameters of the model, we adopt a supervised approach, which requires bounding box annotations for the holistic object as well as body parts of the object. 

During training, we can determine the positions and scales for each node in our model given the ground truth annotations. However, it is difficult to decide whether a node should be switched off or not. The reason is that in our model a node is switched off because it is hard to be detected, which cannot be decided according to the annotations. Note that if a node is not detected, it does not simply mean that the node is invisible or occluded. For example, the holistic body of a highly deformed cat is certainly visible, but is usually hard to detect and we would rather switch it off. Hence, we propose a procedure to assign values to the switch variables $\gamma$ for each positive training example. 

We independently train the appearance models ($\phi$ in Equation \ref{eq:score}) for each node and use their activations to decide whether a body part or holistic object should be switched off or not for a particular training example. Specifically, for node $i$, we set a threshold on its unary term $\phi_i$, and if it is not ``detected'', we label it as switched off in our model, \ie $\gamma_i=0$. We consider a part or holistic object as ``detected'' if there is at least one corresponding activation that has at least 40\% bounding box overlap with it. We use intersection over union (IOU) as the metric of overlap. The thresholds are chosen to ensure that there are few false negatives and a limited number of false positives. 

In this way, we determine the label ${\bf z}$ for each positive training example. Note that our model (Equation~\ref{eq:score}) is linear in the parameter $\boldsymbol\beta = [\boldsymbol{w}, \boldsymbol{b}]$, so we use a linear max-margin learning framework to learn the parameters:
\begin{equation*}
\begin{aligned}
& \min_{\beta, \xi} 
& & \frac{1}{2} \| \boldsymbol\beta \|_2 + C \sum_{i} \xi_i \\
& \text{s.t.} 
& & \boldsymbol\beta \cdot \boldsymbol\Phi_i ({\bf z}) \geq 1 - \xi_i, \forall i \in \text{pos} \\
& & & \boldsymbol\beta \cdot \boldsymbol\Phi_i({\bf z}) \leq -1 + \xi_i, \forall i \in \text{neg}
\end{aligned}
\label{eq:regr}
\end{equation*}
where $\boldsymbol\Phi_i(.)$ is a sparse feature vector representing the $i$-th example and is the concatenation of the appearance, bias, and spatial and scale consistency terms. The above constraints encourage positive examples to be scored higher than 1 (the margin) and the negative examples, which we mine from the negative images using the inference method above, lower than -1. The objective function penalizes violations using slack variables $\xi_i$.

\subsection{Post Processing}
We post-process our results to generate a bounding box for detections whose holistic object node is switched off. Also, we remove multiple detections by a novel part-based non-maximum suppression to get the final output. 

\noindent \textbf{Bounding Box Generation:} For instances with the holistic object detected, we simply use the bounding box for the holistic object  hypothesis as the detection output. For other instances, we use the configuration of their body parts to generate a bounding box for them (since there is no bounding box associated to the holistic object). For this purpose, we learn a mapping function from the bounding boxes of body parts to the upper-left $(x_1, y_1)$ and lower-right $(x_2, y_2)$ corners of the holistic object bounding box.

More specifically, for configurations with $n$ body parts switched on, the object bounding box generation is performed based on the $4n$-dimensional vector $g(z)$, which contains the locations of the upper-left and lower-right corners of the part boxes. We use the annotated holistic object and body part bounding boxes to learn a linear function for estimating $[x_1, y_1, x_2, y_2]$ from $g(z)$. This is done efficiently via least-squares regression, independently for each detectability pattern. 

\noindent \textbf{Part-based Non-Maximum suppression:} Using the inference procedure described above, a single holistic object or body part hypothesis can be used multiple times in different detections. This may produce duplicate detections for the same object. We design a greedy part-based non-maximum suppression to prevent this. There is a score associated to each detection. We sort the detections by their score and start from the highest scoring detection and remove the ones whose holistic object or parts hypotheses are shared with any previously selected detection. After this step, we generate the object bounding box using the above procedure to get the final results.

\section{Experimental Evaluation}
\label{sec:exp}

This section describes our experimental setup, presents a comparative performance evaluation of the proposed method and shows the results of various diagnostic experiments. We report results for the six animal categories in PASCAL VOC 2010 dataset~\cite{pascal-voc-2010}. The considered categories are highly deformable and appear at various scales with different degrees of occlusion. So they serve a suitable testbed for our model. For all of the experiments in the paper, we use \texttt{trainval} subset of PASCAL VOC 2010 detection for training and the \texttt{test} subset for testing and evaluation. 

\begin{figure}
\centering
\includegraphics[width=\linewidth,trim=10 8 10 10,clip=true]{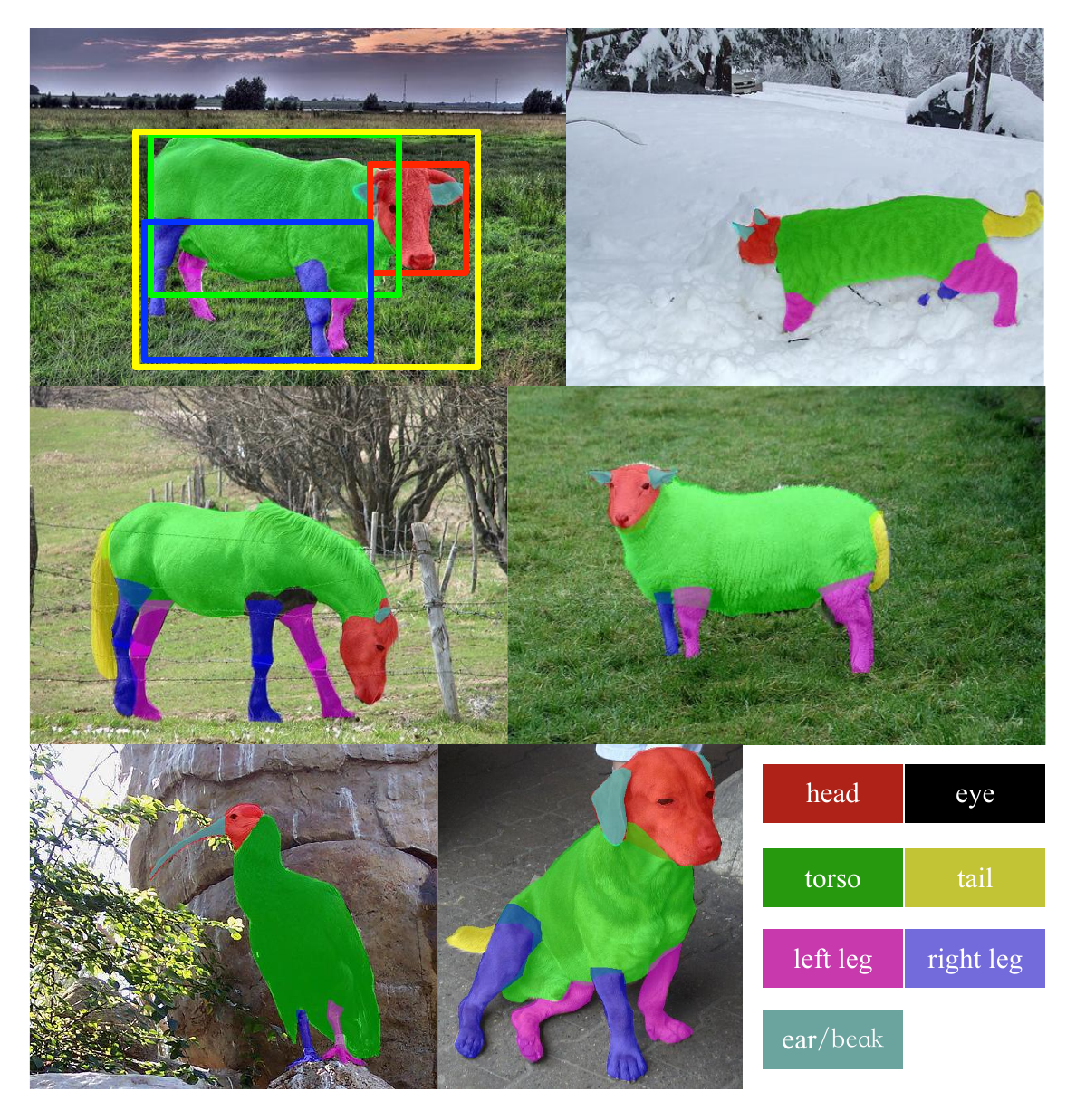}
\caption{{\bf Example annotations of the dataset}. The dataset provides segmentation masks for each body part of the object, but we use the bounding boxes around the masks of head, torso and legs (as shown in the cow image) for training and evaluation.}
\label{fig:dataset}
\end{figure}

\noindent \textbf{Dataset.} We augment PASCAL 2010 dataset with body part annotations as our model is based on body parts. For this purpose, we labeled detailed segmentation masks and bounding boxes of body parts. Figure~\ref{fig:dataset} shows the annotations for some example instances. We note that this is a different annotation than that used in \cite{azizpour12,sun11}. In our experiments, we use the bounding box of the body parts and holistic object for training and evaluation (and not the masks).

\subsection{Implementation detail}
In our experiments, the nodes of our graphical model correspond to the holistic object, head, torso and legs. Note, however,  that our method is not restricted to the types of parts used in this paper and can be easily generalized to other categories and parts.

Our method uses separately trained unary appearance models to prune out the search space and also to infer the switch variables for each positive training example. Unless stated otherwise, we use DPM (\texttt{voc-release5}) as a classifier to obtain the appearance term for our model. Deformable HOG templates seem more robust than rigid HOG templates for representing parts. We use the sigmoid function $\sigma(x)=\frac{1}{1+exp(-1.5x)}$ to renormalize the unary scores.

We adopt the code of LIBLINEAR~\cite{liblinear} to learn our model parameters. We set $C=1$ in all of our experiments. The inference takes around 16 seconds for PASCAL images using a single 2.2 GHz CPU. 

\subsection{Comparison with other methods}

Table~\ref{tab:fullres} shows the results of our model with and without the holistic object node. We also provide a comparison with the DPM~\cite{felzenswalb10} and strongly supervised methods such as Poselets~\cite{Bourdev09}, \cite{azizpour12}, \cite{parkhi11}, and \cite{gu12}. These methods also use additional annotations such as keypoints, bounding boxes for semantic regions or segmentation masks. In the table, we show the results of our method using two different classifiers (\cite{felzenswalb10} and \cite{fidler13}) for obtaining the unary appearance terms for the holistic object node.

\begin{table}
\scriptsize
\centering
\rowcolors{2}{}{gray!35}
\addtolength{\tabcolsep}{-0.3pt}
\begin{tabular}{ l c c c c c c c }
\toprule[0.2 em] %
& Bird & Cat & Cow & Dog & Horse & Sheep & mAP \\
Ours w. segDPM & \textbf{26.7} & \textbf{52.4} & \textbf{36.2} & \textbf{42.7} & \textbf{53.0} & \textbf{37.4} & \textbf{41.4}\\
Ours w. DPM & 15.6 & 37.3 & 30.3 & 31.3 & 49.0 & 30.1 & 32.3 \\
Ours w/o holistic & 11.0 & 35.7 & 25.3 & 27.2 & 40.7 & 23.9 & 27.3 \\

\midrule
DPM~\cite{felzenswalb10} & 11.0 & 23.6 & 23.2 & 20.5 & 42.5 & 29.0 & 25.0\\
Poselets~\cite{Bourdev09}& 8.5 & 22.2 & 20.6 & 18.5 & 48.2 & 28.0 & 24.3\\
Sup-DPM~\cite{azizpour12} & 11.3 & 27.2 & 25.8 & 23.7 & 46.1 & 28.0 & 27.0\\
M-Comp~\cite{gu12} & 18.1 & 45.4 & 23.4 & 35.2 & 40.7 & 34.6 & 32.9 \\
DisPM~\cite{parkhi11} & - & 45.3 & - & 36.8 & - & - & -\\
segDPM~\cite{fidler13}  & 25.3 & 48.8 & 30.4 & 37.7 & 46.0 & 35.7 & 37.3\\
\bottomrule[0.1 em]
\end{tabular}
\vspace{0.2mm}
\caption{Average precision for detection of animals of PASCAL VOC 2010. The best results are obtained by our method using segDPM for the holistic object and DPM for the body parts. Also note that our method outperforms DPM, and Sup-DPM, even if we use only the body parts (the third row).}
\label{tab:fullres}
\end{table}

\begin{figure*}
\centering
  \includegraphics[width=\linewidth,trim=5 0 8 0,clip=true]{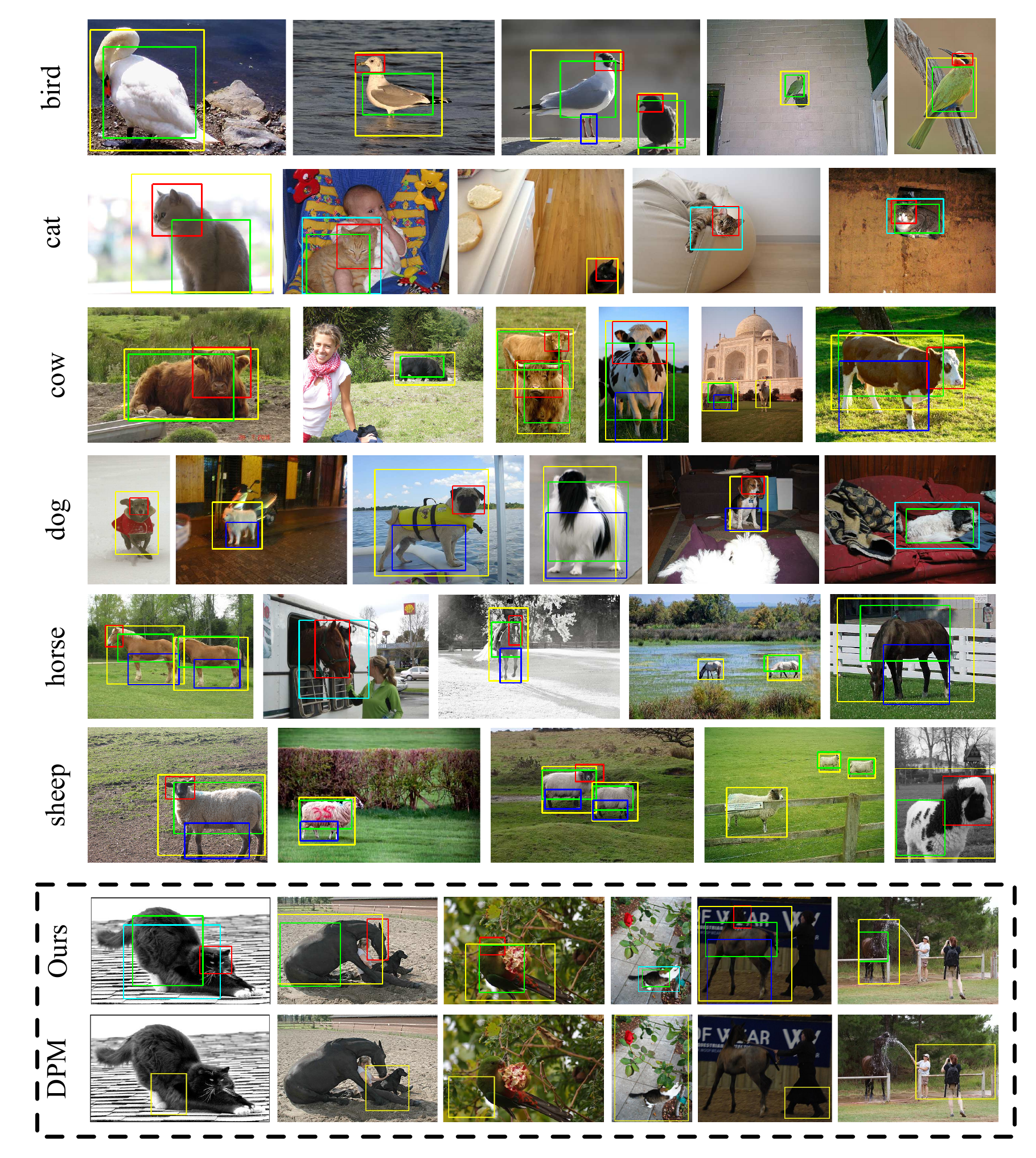}
  \caption{Our method provides a richer representation for the deformable objects and improves detection results on difficult examples. The red, green and blue boxes correspond to head, torso and legs, respectively. The holistic object boxes are shown in yellow, and the cyan boxes correspond to the generated holistic object bounding box (note that we should generate a bounding box for patterns whose holistic object node is switched off). In the dashed box, we also show examples of animals correctly localized by our method but missed by DPM. }
  \label{fig:lastres}
\end{figure*}

As shown in the table, our model without the holistic object node already achieves competitive results. Adding the holistic object node further boosts the performance to 32.3\% AP, which is 7.3\% AP higher than DPM performance. In addition, we provide richer description for objects since we output a bounding box for body parts as well. Figure~\ref{fig:lastres} shows examples of our object detection results, and also some typical examples that are correctly localized by our method but missed by DPM.

Our model with segDPM unary for the holistic object clearly outperforms all of the previous methods by a large margin and provides 4.1\% AP improvement over the state-of-the-art~\cite{fidler13}. The high performance of our method is not only due to using a strong unary term as we outperform the unary alone by a large margin. Since segDPM~\cite{fidler13} is not designed for object parts, we use it only for the holistic object node. Our model with DPM unary is on par with \cite{gu12}. Nevertheless, in terms of memory and computation, our method is about an order of magnitude more efficient since \cite{gu12} relies on the output of a segmentation method, and uses multiple features, spatial pyramids and kernels. Our model is much simpler as it uses only HOG feature and a few linear classifiers.

\subsection{Diagnostic Experiments}
\label{sec:diagnose}
We proposed a flexible model to better handle large deformations, occlusions and low resolution. Also, our model is able to provide a richer description than bounding box for objects. In this section, we perform a set of experiments to support these claims. 

\subsubsection{Importance of switch variables}
Our model allows some of the nodes to be switched off. We perform an additional experiment to show that this is critical for high performance. We enforce every node in our model to be switched on and perform experiments both with and without the holistic object node. Table~\ref{tab:switch} shows the performance degrades drastically if the nodes cannot be switched off, which confirms that adaptively switching off some of the nodes in the model is important. 

\begin{table}
\scriptsize
\centering
\rowcolors{2}{}{gray!35}
\addtolength{\tabcolsep}{-0.6pt}
\begin{tabular}{ l c c c c c c c }
\toprule[0.2 em] %
& Bird & Cat & Cow & Dog & Horse & Sheep & mAP \\
NSO (w/o holistic) & 0.4 & 21.0 & 8.1 & 22.4 & 26.6 & 14.0 & 15.4 \\
NSO & 0.6 & 20.8 & 8.8 & 24.4 & 28.3 & 16.7 & 16.6 \\
Ours w. DPM& \textbf{15.6} & \textbf{37.3} & \textbf{30.3} & \textbf{31.3} & \textbf{49.0} & \textbf{30.1} & \textbf{32.3} \\
\bottomrule[0.1 em]
\end{tabular}
\vspace{0.1mm}
\caption{Average precision for detection of animals of PASCAL VOC 2010, when we enforce all the nodes to be switched on.  Performance degrades drastically if the nodes cannot be switched off. NSO stands for No Switch Off.}
\label{tab:switch}
\end{table}

\subsubsection{Small scale objects} 
In this experiment we show that the holistic-only detectability pattern (where all body parts are switched off) better captures objects at low resolution (\ie small objects). 

We set a threshold to get a high recall and divide the recalled instances of each object category into multiple size classes. We follow the same division as Hoiem \etal~\cite{hoiem12}, where each instance is assigned to a size category, depending on its percentile size: extra-small ({\bf XS}: bottom 10\%); small ({\bf S}: next 20\%); medium ({\bf M}: next 40\%); large ({\bf L}: next 20\%); extra-large ({\bf XL}: next 10\%). We then compute what percentage of each size class has been inferred as holistic-only detectability pattern.

As shown in Table~\ref{tab:size}, the holistic-only detectability pattern has the highest rate for the instances of the extra-small class and it describes fewer instances of the other sizes. This follows our argument that typically it is hard to detect body parts for tiny objects, and a model without body parts seems more suitable for the instances of low resolution. This is also observed by \cite{park10} for the task of pedestrian detection.
\begin{table}[t!]
\centering
\addtolength{\tabcolsep}{0.5pt}
\begin{tabular}{|l||c|c|c|c|c|}
\hline
     & \textbf{XS} & \textbf{S} & \textbf{M} & \textbf{L} & \textbf{XL} \\
\hline\hline     
Bird & 66.7\% & 38.0\% & 24.0\% & 29.3\% & 21.7\% \\
\hline
Cat  & 22.2\% & 3.9\% &3.9\% & 1.7\% & 1.1\% \\
\hline
Cow  & 66.7\% & 29.1\% & 15.6\% & 12.7\% & 28.6\% \\
\hline
Dog  & 27.8\% & 7.8\% & 8.7\% & 6.0\% & 4.6\% \\
\hline
Horse& 52.2\% & 18.1\% & 11.7\% & 13.8\% & 19.1\% \\
\hline
Sheep& 52.5\% & 22.0\% & 11.6\% & 14.6\% & 12.2\% \\
\hline
\end{tabular}
\vspace{0.6mm}
\caption{Percentage of instances described by the holistic-only detectability pattern. We divide the instances of a category into 5 size classes and show the ratio of instances of each class that are inferred as holistic-only detectability pattern. The holistic-only detectability pattern is more effective for the XS class.}
\label{tab:size}
\vspace{-1mm}
\end{table}

\subsubsection{Part localization} 
Our method can localize object parts and provides a richer description of objects. We adopt the widely used measure of PCP (Percentage of Correctly estimated body Parts)~\cite{Ferrari08} to evaluate the object parts localized by our method. Following~\cite{yang11}, for each ground truth object, we consider the detection with the highest score that has more than 50\% overlap with its bounding box. This factors out the effect of the detection. A part is considered as correctly localized if it has more than 40\% overlap with the ground truth annotation. Table~\ref{tab:part_loc} shows PCP for parts used in our experiments. Our model outputs fairly precise locations for body parts. 

Since our part detection does not always activate for every detected object, we also show the Percentage of Objects with the Part estimated (POP) in Table~\ref{tab:part_loc}, where we show the percentage of objects that have a detection for a certain part type. We can see some interesting patterns for different kinds of animals. As expected, head and torso are used more often than legs, which is probably because legs are prone to be truncated or occluded and are sometimes tiny (\eg bird's legs). Head is used very often for cat and dog, while for bird and sheep, torso seems more reliable. The reason is probably that head is a distinctive part for cat and dog as it is observed by~\cite{parkhi11}, but for bird and sheep, it is often too small. 

\begin{table}[t!]
\centering
\addtolength{\tabcolsep}{4.0pt}
\begin{tabular}{|l||c|c|c|c|}
\hline
     & \textbf{Head} & \textbf{Torso} & \textbf{Legs} \\
\hline\hline     
Bird & 27.0 / 61.8 & 62.9 / 83.2 & 2.4 / 18.2 \\
\hline
Cat  & 73.5 / 77.3 & 67.1 / 71.4 & 18.5 / 28.1 \\
\hline
Cow  & 36.1 / 88.9 & 64.2 / 89.2 & 19.7 / 92.6 \\
\hline
Dog  & 67.7 / 75.0 & 51.7 / 57.9 & 29.7 / 44.9 \\
\hline
Horse& 52.0 / 66.8 & 71.2 / 92.5 & 37.7 / 82.5 \\
\hline
Sheep& 24.9 / 70.6 & 79.2 / 88.6 & 16.1 / 80.3 \\
\hline
\end{tabular}
\vspace{0.5mm}
\caption{Part localization performance on PASCAL VOC 2010. Reported numbers are POP / PCP, where POP stands for Percentage of Objects with the Part estimated, and PCP stands for Percentage of Correctly estimated body Parts (see text for details).}
\label{tab:part_loc}
\end{table}

\section{Conclusion}
\label{sec:conc}
In this paper, we propose a new part-based model to better detect and describe objects. Our model allows each body part or the the holistic object node to be switched off to better handle highly deformed, occluded or low resolution objects. To detect objects we perform inference on a loopy graph that directly models the spatial and scale relationships between all pairs of body parts and the holistic object. We compare our method with other strongly supervised methods and also the state-of-the-art on PASCAL VOC 2010 dataset and show a 4.1\% AP improvement over the state-of-the-art~\cite{fidler13} for animal categories. We also outperform DPM~\cite{felzenswalb10} and Sup-DPM~\cite{azizpour12} even if we simplify our model by using only the body parts (\ie no holistic model). We also show that our model localizes the body parts fairly reliably. To train and evaluate our model, we provide detailed mask annotations for body parts of the six animal categories on PASCAL VOC 2010.  
\\
\\
\noindent {\bf Acknowledgments}~{This work has been supported by grants ARO 62250-CS, N00014-12-1-0883, and the Center for Minds, Brains and Machines (CBMM), funded by NSF STC award CCF-1231216.}

{\small
\bibliographystyle{ieee}
\bibliography{biblio}
}


\end{document}